# An Efficient Message-Passing Algorithm for the M-Best MAP Problem

**Dhruv Batra**
TTI-Chicago
dbatra@ttic.edu

## Abstract

Much effort has been directed at algorithms for obtaining the highest probability configuration in a probabilistic random field model – known as the maximum a posteriori (MAP) inference problem. In many situations, one could benefit from having not just a single solution, but the top $M$ most probable solutions – known as the M-Best MAP problem.

In this paper, we propose an efficient message-passing based algorithm for solving the M-Best MAP problem. Specifically, our algorithm solves the recently proposed Linear Programming (LP) formulation of M-Best MAP [7], while being *orders of magnitude* faster than a generic LP-solver. Our approach relies on studying a particular partial Lagrangian relaxation of the M-Best MAP LP which exposes a natural combinatorial structure of the problem that we exploit.

## 1 Introduction

A large number of problems in computer vision, natural language processing and computational biology can be formulated as the search for the most probable state under a discrete probabilistic graphical model – known the MAP inference problem.

In a number of such applications, one can benefit from having not just a single best solution, rather a list of M-best hypotheses. For example, sentences are often ambiguous and machine translation systems benefit from working with multiple plausible parses of a sentence. In computational biology, practitioners are often interested in computing the top M most stable configurations of a protein structure. Moreover, computing such a set of M-best hypotheses is useful for assessing the sensitivity of the model w.r.t. variations in the input and/or the parameters of the model.

In the graphical models literature, this problem is known as the M-Best MAP problem [7, 18, 28]. Interestingly (and perhaps understandably), algorithms for the M-Best MAP problem have closely followed the development of the algorithms for solving the MAP problem. Similar to MAP, the first family of algorithms for M-Best MAP [18, 20] were junction-tree based exact algorithms, feasible only for low-treewidth graphs. For high-treewidth models, where Belief Propagation (BP) is typically used to perform approximate MAP inference, Yanover and Weiss [28] showed how the pseudo-max-marginals produced by BP may be used to compute approximate M-Best MAPs.

However, with the development of Linear Programming (LP) relaxations for MAP, this concurrence between MAP and M-Best MAP is no longer true. While message-passing algorithms for solving the MAP LP were available as soon as the LPs were studied [8, 13, 19, 24, 26], no such algorithm is known for solving the M-Best MAP LP [7]. This discrepancy is not merely a theoretical concern – large-scale empirical comparisons [27] have found that message-passing algorithms for MAP significantly outperform commercial LP solvers (like CPLEX). More importantly, message-passing algorithms can be applied to large-scale problems where solvers like CPLEX simply would not scale. Thus, if we are to apply M-Best MAP to real instances appearing in computational biology, computer vision and NLP, we must develop scalable distributed message-passing algorithms.

**Overview.** The principal contribution of this paper is to develop an efficient message-passing algorithm for the M-Best MAP problem in discrete undirected graphical models, specifically Markov Random Fields (MRFs). Our approach studies a particular partial Lagrangian relaxation of the M-Best MAP LP [7] which exposes the natural modular structure in the problem. For graphs with cycles, this Lagrangian relaxation involves an exponentially large set of dual variables, and we use a dynamic subgradient method (DSM) [4] for solving this Lagrangian dual. DSMs are a recently formalized class of methods that interleave a separation oracle procedure (that selects a subset of active dual variables), with the dual update procedure (that takes a step in the direction of the subgradient).

At a high-level, our algorithm brings MAP and M-Best MAP to an equal footing vis-a-vis a message-passing algorithm for solving the corresponding LP-relaxation. Importantly, our algorithm retains all the guarantees of the LP formulation of Fromer and Globerson [7], while being *orders of magnitude* faster. Similar to the observations in [27] for MAP, we find that our algorithm enables solving M-Best MAP on large instances that were unsolvable with generic LP solvers.

**Outline.** We begin with a brief history of the M-Best MAP problem in Section 2; present preliminaries and background in Section 3; revisit the M-Best MAP LP formulation of Fromer and Globerson [7] in Section 4; study the Lagrangian dual of this LP and present a message-passing dual ascent algorithm for tree-MRF in Section 5 and for general MRFs in Section 6.

## 2  Previous Work on M-Best MAP

The problem of finding the top M solutions to a general combinatorial optimization problem (not just inference in MRFs) has typically been studied in the context of k-shortest paths [6] in a search graph. Lawler [16] proposed a general algorithm to compute the top M solutions for a large family of discrete optimization problems, and the ideas used in Lawler's algorithm form the basis of most algorithms for the M-Best MAP problem. If the complexity of finding the best solution is $T(n)$, where $n$ is the number of variables, Lawler's algorithm solves $n$ new problems. The best solution among these $n$ problems is the second best solution to the original problem. Thus at an $O(nM)$ multiplicative overhead, the top M solutions can be iteratively found. Hamacher and Queyranne [10] reduced this overhead to $O(M)$ by assuming access to an algorithm that can compute the first and the second best solutions.

Dechter and colleagues [5, 17] have recently provided dynamic-programming algorithms for M-Best MAP, but these are exponential in treewidth. Yanover and Weiss [28] proposed an algorithm that requires access only to max-marginals. Thus, for certain classes of MRFs that allow efficient exact computation of max-marginals, *e.g.* binary pairwise supermodular MRFs [11], M-Best solutions can be found for arbitrary treewidth graphs. Moreover, *approximate* M-Best solutions may be found by approximating the max-marginal computation, *e.g.* via loopy BP.

More recently, Fromer and Globerson [7] provided a LP view of the M-Best MAP problem. They proposed an algorithm (STRIPES) that repeatedly partitions the space of solutions and solves a $2^{nd}$-Best MAP LP within each partition (with a generic LP-solver). We revisit [7] in detail in Section 4, and describe a significantly more efficient message-passing algorithm for solving the LP in the remainder of the paper.

## 3  Preliminaries: MAP-MRF Inference

**Notation.** For any positive integer $n$, let $[n]$ be shorthand for the set $\{1, 2, \ldots, n\}$. We consider a set of discrete random variables $\mathbf{x} = \{x_i \mid i \in [n]\}$, each taking value in a finite label set, $x_i \in X_i$. For a set $A \subseteq [n]$, we use $x_A$ to denote the tuple $\{x_i \mid i \in A\}$, and $X_A$ to be the cartesian product of the individual label spaces $\times_{i \in A} X_i$. For ease of notation, we use $x_{ij}$ as a shorthand for $x_{\{i,j\}}$. For two vector $\boldsymbol{a}, \boldsymbol{b} \in \mathbb{R}^d$, we use $\boldsymbol{a} \cdot \boldsymbol{b}$ to denote the inner product.

**MAP.** Let $G = (\mathcal{V}, \mathcal{E})$ be a graph defined over these variables, *i.e.* $\mathcal{V} = [n]$, $\mathcal{E} \subseteq \binom{\mathcal{V}}{2}$, and let $\theta_A : X_A \to \mathbb{R}$, $(\forall A \in \mathcal{V} \cup \mathcal{E})$ be functions defining the energy at each node and edge for the labeling of variables in scope. The goal of MAP inference is to find the labeling $\mathbf{x}$ of the variables that minimizes this real-valued energy function:

$$\min_{\mathbf{x} \in X_\mathcal{V}} \sum_{A \in \mathcal{V} \cup \mathcal{E}} \theta_A(x_A) \tag{1a}$$

$$= \min_{\mathbf{x} \in X_\mathcal{V}} \sum_{i \in \mathcal{V}} \theta_i(x_i) + \sum_{(i,j) \in \mathcal{E}} \theta_{ij}(x_i, x_j). \tag{1b}$$

The techniques developed in this paper are naturally applicable to higher-order MRFs as well. However, to simplify the exposition we restrict ourselves to pairwise energy functions.

**MAP Integer Program.** MAP inference is typically set up as an integer programming problem over boolean variables. For each node and edge $A \in \mathcal{V} \cup \mathcal{E}$, let $\boldsymbol{\mu}_A = \{\mu_A(s) \mid s \in X_A, \mu_A(s) \in \{0,1\}\}$, be a vector of indicator variables encoding all possible configurations of $x_A$. If $\mu_A(s)$ is set to 1, this implies that $x_A$ takes label $s$. Moreover, let $\boldsymbol{\theta}_A = \{\theta_A(s) \mid s \in X_A\}$ be a vector holding energies for all possible configurations of $x_A$, and $\boldsymbol{\mu} = \{\boldsymbol{\mu}_A \mid A \in \mathcal{V} \cup \mathcal{E}\}$ be a vector holding the entire configuration. Using this notation, the MAP inference integer program can be written as:

$$\min_{\boldsymbol{\mu}_i, \boldsymbol{\mu}_{ij}} \sum_{i \in \mathcal{V}} \boldsymbol{\theta}_i \cdot \boldsymbol{\mu}_i + \sum_{(i,j) \in \mathcal{E}} \boldsymbol{\theta}_{ij} \cdot \boldsymbol{\mu}_{ij} \tag{2a}$$

$$s.t. \sum_{s \in X_i} \mu_i(s) = 1 \qquad \forall i \in \mathcal{V} \tag{2b}$$

$$\sum_{s \in X_i} \mu_{ij}(s,t) = \mu_j(t) \qquad \forall \{i,j\} \in \mathcal{E} \tag{2c}$$

$$\sum_{t \in X_j} \mu_{ij}(s,t) = \mu_i(s) \qquad \forall \{i,j\} \in \mathcal{E} \tag{2d}$$

$$\mu_i(s), \mu_{ij}(s,t) \in \{0,1\} \quad \forall i \in \mathcal{V}, \forall \{i,j\} \in \mathcal{E} \tag{2e}$$

Here (2b) enforces that exactly one label is assigned to a variable, and (2c),(2d) that assignments are consistent across edges. To be concise, we will use $\mathcal{P}(G)$ to denote the set of $\boldsymbol{\mu}$ that satisfy the three constraints (2b),(2c),(2d). Thus,

the above problem (2) can be written concisely as:
$$\min_{\boldsymbol{\mu} \in \mathcal{P}(G),\, \mu_A(s) \in \{0,1\}} \sum_{A \in \mathcal{V} \cup \mathcal{E}} \boldsymbol{\theta}_A \cdot \boldsymbol{\mu}_A.$$

**MAP LP.** Problem (2) is known to be NP-hard in general [21]. A number of techniques [8, 13, 19, 24, 26] solve a Linear Programming (LP) relaxation of this problem, which is given by relaxing the boolean constraints (2e) to the unit interval, i.e. $\mu_i(s), \mu_{ij}(s,t) \geq 0$. Thus, the MAP LP minimizes the energy over the following polytope: $\mathcal{L}(G) = \{\mu_A(\cdot) \geq 0, \mid \boldsymbol{\mu} \in \mathcal{P}(G)\}$, also known as the *local polytope*. The LP relaxation of MAP is known to be *tight* for special cases like tree-graphs and binary submodular energies [25], meaning that the optimal vertex of the local polytope is an integer $\boldsymbol{\mu}_A \in \{0, 1\}$, for these special cases.

## 4 M-Best MAP Linear Program

Let us now revisit the M-Best MAP LP formulation [7]. Let $\boldsymbol{\mu}^m$ denote the $m^{th}$-best MAP. Thus $\boldsymbol{\mu}^1$ is the MAP, $\boldsymbol{\mu}^2$ is the second-best MAP and so on. The M-Best MAP integer program is given by

$$\boldsymbol{\mu}^M = \operatorname*{argmin}_{\boldsymbol{\mu} \in \mathcal{P}(G),\, \mu_A(s) \in \{0,1\}} \sum_{A \in \mathcal{V} \cup \mathcal{E}} \boldsymbol{\theta}_A \cdot \boldsymbol{\mu}_A \quad (3a)$$

$$s.t. \quad \boldsymbol{\mu} \neq \boldsymbol{\mu}^m \quad \forall m \in [M-1] \quad (3b)$$

While the MAP integer program suggested a natural LP relaxation, that is not the case with the M-Best MAP integer program due to the exclusion constraints (3b), which are not linear constraints. Fromer and Globerson [7] introduced a concise representation of a polytope called *assignment-excluding local polytope* $\mathcal{L}(G, \{\boldsymbol{\mu}^m\}_1^{M-1})$ that excludes previous solutions $\{\boldsymbol{\mu}^m \mid m \in [M-1]\}$ with the help of additional linear inequalities, called the *spanning-tree inequalities*.

**Spanning Tree Inequalities.** Let $T \subseteq \mathcal{E}$ be a spanning tree of $G$, and $\mathcal{T}(G)$ be the set of all such spanning trees in $G$. Let $d_i^T$ be the degree of node $i$ in $T$. Let us define $I^T(\boldsymbol{\mu}, \boldsymbol{\mu}^m) \triangleq \sum_{i \in \mathcal{V}} (1 - d_i^T) \boldsymbol{\mu}_i \cdot \boldsymbol{\mu}_i^m + \sum_{(i,j) \in T} \boldsymbol{\mu}_{ij} \cdot \boldsymbol{\mu}_{ij}^m.$

Now, a spanning tree inequality is defined as: $I^T(\boldsymbol{\mu}, \boldsymbol{\mu}^m) \leq 0$.

Notice that $I^T(\boldsymbol{\mu}^m, \boldsymbol{\mu}^m) = 1$. Moreover, it can be shown [7] that $I^T(\boldsymbol{\mu}, \boldsymbol{\mu}^m) \leq 0,\, \forall \boldsymbol{\mu} \neq \boldsymbol{\mu}^m$. Thus, the spanning tree inequality *separates* the vertex $\boldsymbol{\mu}^m$ from other vertices in the polytope.

**Assignment-Excluding Local Polytope (AELP)** is defined as:

$$\mathcal{L}(G, \{\boldsymbol{\mu}^m\}_1^{M-1}) = \Big\{\boldsymbol{\mu} \;\Big|\; \boldsymbol{\mu} \in \mathcal{L}(G),\ I^T(\boldsymbol{\mu}, \boldsymbol{\mu}^m) \leq 0,$$
$$\forall T \in \mathcal{T}(G), \forall m \in [M-1]\Big\}. \quad (4)$$

Thus, we can see that the AELP excludes each of the previous solutions $\{\boldsymbol{\mu}^m \mid m \in [M-1]\}$ with the help of spanning tree inequalities.

Recall that the LP relaxation over the local polytope for a tree-structured MRFs is tight. It can also be shown [7] that the LP relaxation over AELP for tree MRFs is tight for $m = 2$. However, for $m \geq 3$ in tree MRFs and any for any $m \geq 1$ in loopy MRFs, the AELP is not guaranteed to be a tight relaxation.

**Efficient Separation Oracle.** Note that for tree MRFs, there is a single spanning tree inequality for each previous solution since $|\mathcal{T}(G)| = 1$, while for general graphs there may be an exponentially large collection of spanning trees, e.g. $|\mathcal{T}(G)| = n^{n-2}$ for complete graphs. However, not all such inequalities need to be explicitly included. We can use a cutting-plane algorithm that maintains a working set of spanning trees $\mathcal{T}'$ and incrementally adds the most violated inequality: $\mathcal{T}' \longleftarrow \mathcal{T}' \cup \operatorname{argmax}_{T \in \mathcal{T}(G)} I^T(\boldsymbol{\mu}, \boldsymbol{\mu}^m)$. For a given $\boldsymbol{\mu}$, Fromer and Globerson [7] showed that this separation oracle can be efficiently implemented with a maximum-weight spanning tree algorithm with the edge weights given by $w_{ij} = \boldsymbol{\mu}_{ij} \cdot \boldsymbol{\mu}_{ij}^m - \boldsymbol{\mu}_i \cdot \boldsymbol{\mu}_i^m - \boldsymbol{\mu}_j \cdot \boldsymbol{\mu}_j^m$.

Notice that this algorithm requires solving a linear program over the AELP in each iteration. For large problems arising in computer vision and computational biology, solving this LP with a standard LP-solver *even once* may be infeasible. In the next section, we present our proposed message-passing algorithm for solving the M-Best MAP LP.

## 5 M-Best MAP Lagrangian Relaxation: Tree-MRF

Let us first restrict our attention to tree-structured MRFs. This is simple enough a scenario to describe the main elements of our approach; we then discuss the general case in Section 6. For tree MRFs, there is a single spanning tree inequality (for each $m$), and to simplify notation we will refer to $I^T(\boldsymbol{\mu}, \boldsymbol{\mu}^m)$ simply as $I(\boldsymbol{\mu}, \boldsymbol{\mu}^m)$. Then M-Best MAP LP can be written as:

$$\min_{\boldsymbol{\mu} \in \mathcal{L}(G)} \sum_{A \in \mathcal{V} \cup \mathcal{E}} \boldsymbol{\theta}_A \cdot \boldsymbol{\mu}_A \quad (5a)$$

$$s.t. \quad I(\boldsymbol{\mu}, \boldsymbol{\mu}^m) \leq 0 \quad \forall m \in [M-1] \quad (5b)$$

Now instead of solving the above problem in the primal with an LP-solver as Fromer and Globerson [7] did, we will study the Lagrangian relaxation of this LP, formed by dualizing the spanning tree constraints:

$$f(\boldsymbol{\lambda}) = \min_{\boldsymbol{\mu} \in \mathcal{L}(G)} \sum_{A \in \mathcal{V} \cup \mathcal{E}} \boldsymbol{\theta}_A \cdot \boldsymbol{\mu}_A + \sum_{m=1}^{M-1} \lambda_m I(\boldsymbol{\mu}, \boldsymbol{\mu}^m), \quad (6)$$

where $\boldsymbol{\lambda} = \{\lambda_m \mid m \in [M-1]\}$ is the set of Lagrange multipliers, that determine the weight of the penalty imposed for violating the spanning tree constraints.

Note that this is a partial Lagrangian because we have only dualized the spanning tree constraints (5b), and have not dualized the constraints hidden inside the local polytope $\mathcal{L}(G)$.

**Key Idea: Exploiting Structure.** The partial Lagrangian immediately exposes a structure in the problem that the primal formulation was obfuscating, namely that the spanning tree inequality distributes according to a tree structure, *i.e.*

$$f(\boldsymbol{\lambda}) = \min_{\boldsymbol{\mu} \in \mathcal{L}(G)} \left\{ \sum_{i \in \mathcal{V}} \left( \boldsymbol{\theta}_i + \sum_{m=1}^{M-1} \lambda_m (1-d_i) \boldsymbol{\mu}_i^m \right) \cdot \boldsymbol{\mu}_i \right. \\ \left. + \sum_{(i,j) \in \mathcal{E}} \left( \boldsymbol{\theta}_{ij} + \sum_{m=1}^{M-1} \lambda_m \boldsymbol{\mu}_{ij}^m \right) \cdot \boldsymbol{\mu}_{ij} \right\}. \quad (7)$$

Also recall that for a tree the local polytope $\mathcal{L}(G)$ has integral vertices. Thus the minimization above can be efficiently performed by running a combinatorial optimization algorithm (the standard two-pass max-product BP) on this perturbed MRF, and does not need to be solved with a generic LP solver. As we will see next, being able to efficiently *evaluate* the Lagrangian is all we need to be able to *optimize* the Lagrangian dual.

### 5.1 Projected Supergradient Ascent on the Lagrangian Dual

From the theory of Lagrangian duality, we know that for all values of $\boldsymbol{\lambda} \geq 0$, $f(\boldsymbol{\lambda})$ is a lower-bound on the value of the primal problem (5). The tightest lower-bound is obtained by solving the Lagrangian dual problem: $\max_{\boldsymbol{\lambda} \geq 0} f(\boldsymbol{\lambda})$. Since $f$ is a non-smooth concave function, this can be achieved by the *supergradient ascent* algorithm, analogous to the subgradient descent for minimizing non-smooth convex functions [22]. Since $\boldsymbol{\lambda}$ is a constrained variable, we follow the projected supergradient ascent algorithm: iteratively updating the Lagrange multipliers according to the following update rule: $\boldsymbol{\lambda}^{(t+1)} \longleftarrow \left[ \boldsymbol{\lambda}^{(t)} + \alpha_t \nabla f(\boldsymbol{\lambda}^{(t)}) \right]_+$, where $\nabla f(\boldsymbol{\lambda}^{(t)})$ is the supergradient of $f$ at $\boldsymbol{\lambda}^{(t)}$, $\alpha_t$ is the step-size and $[\cdot]_+$ is the projection operator that projects a vector onto the positive orthant. If the sequence of multipliers $\{\alpha_t\}$ satisfies $\alpha_t \geq 0$, $\lim_{t \to \infty} \alpha_t = 0$, $\sum_{t=0}^{\infty} \alpha_t = \infty$, then projected supergradient ascent converges to the optimum of the Lagrangian dual [22].

To find the supergradient of $f(\boldsymbol{\lambda})$, consider the following lemma (proved in [14]):

**Lemma 1** *If $g(\boldsymbol{\lambda})$ is a point-wise minimum of linear functions:* i.e. $g(\boldsymbol{\lambda}) = \min_{\boldsymbol{\mu}} \boldsymbol{a}_{\boldsymbol{\mu}} \cdot \boldsymbol{\lambda} + b_{\boldsymbol{\mu}}$, *then one supergradient of $g$ is given by $\nabla g(\boldsymbol{\lambda}) = \boldsymbol{a}_{\hat{\boldsymbol{\mu}}(\boldsymbol{\lambda})}$, where $\hat{\boldsymbol{\mu}}(\boldsymbol{\lambda}) \in \arg\min_{\boldsymbol{\mu}} \boldsymbol{a}_{\boldsymbol{\mu}} \cdot \boldsymbol{\lambda} + b_{\boldsymbol{\mu}}$.*

Notice that $f$ is indeed a point-wise minimum of linear functions. Mapping this lemma to (5), we can see that the supergradient of $f$ for our formulation is given by:

$$\nabla f(\boldsymbol{\lambda}) = \left[ I(\hat{\boldsymbol{\mu}}(\boldsymbol{\lambda}), \boldsymbol{\mu}^1), \ldots, I(\hat{\boldsymbol{\mu}}(\boldsymbol{\lambda}), \boldsymbol{\mu}^{M-1}) \right]^T \quad (8)$$

where $\hat{\boldsymbol{\mu}}(\boldsymbol{\lambda})$ is an optimal primal solution of (5) for the current setting of $\boldsymbol{\lambda}$. Thus the computation of the supergradient can be done with the same dynamic programming algorithm as for evaluating the Lagrangian.

This supergradient (and the update procedure) has an intuitive interpretation. Recall that the Lagrangian relaxation minimizes a linear combination of the energy and the value of the spanning tree inequality, with the weighting given by $\boldsymbol{\lambda}$. If $\hat{\boldsymbol{\mu}}(\boldsymbol{\lambda}^{(t)})$ violates one of the spanning tree constraints, *i.e.* is not different from a previous solution $\boldsymbol{\mu}^m$, then the supergradient w.r.t. $\lambda_m^{(t)}$ will be positive and the cost for violating the constraint will increase after the update, thus encouraging the next solution $\hat{\boldsymbol{\mu}}(\boldsymbol{\lambda}^{(t+1)})$ to satisfy the spanning tree constraints. Conversely, if the constraints are (strictly) satisfied, the supergradient is negative indicating that $\lambda_m^{(t)}$ may be over-penalizing for violations and may be reduced to allow lower energy solutions.

**Tightness of the Lagrangian Relaxation.** The primal problem (5) is an LP, and strong duality holds. Thus, the projected supergradient algorithm described above exactly solves the M-Best MAP LP of Fromer and Globerson [7]. The total complexity of the algorithm is $O(knL^2)$, where $k$ is the number of dual ascent iterations, $n$ is the number of nodes and $L$ is the largest label space, *i.e.* $L = \max_i |X_i|$.

## 6 M-Best MAP Lagrangian Relaxation: General MRFs

In this section, we build on the basic ideas from the previous section to develop an algorithm for general graphs, which may contain exponentially many spanning trees. Recall that the M-Best MAP LP for general graphs is given by:

$$\min_{\boldsymbol{\mu} \in \mathcal{L}(G)} \sum_{A \in \mathcal{V} \cup \mathcal{E}} \boldsymbol{\theta}_A \cdot \boldsymbol{\mu}_A \quad (9a)$$

$$s.t. \quad I^T(\boldsymbol{\mu}, \boldsymbol{\mu}^m) \leq 0, \quad \forall T \in \mathcal{T}(G), \forall m \in [M-1] \quad (9b)$$

where $\mathcal{T}(G)$ is the set of all spanning trees in $G$. As before, the Lagrangian formed by dualizing the spanning tree constraints is given by:

$$f(\boldsymbol{\lambda}) = \min_{\boldsymbol{\mu} \in \mathcal{L}(G)} \sum_{A \in \mathcal{V} \cup \mathcal{E}} \boldsymbol{\theta}_A \cdot \boldsymbol{\mu}_A + \sum_{m=1}^{M-1} \sum_{T \in \mathcal{T}(G)} \lambda_{mT} I^T(\boldsymbol{\mu}, \boldsymbol{\mu}^m) \quad (10)$$

There are two main concerns that prevent us from directly solving this Lagrangian relaxation as before. First, the set of Lagrange multipliers $\boldsymbol{\lambda} = \{\lambda_{mT} \mid m \in [M-1], T \in \mathcal{T}(G)\}$ is exponentially large. And second, the graph is no longer a tree, so the supergradient can no longer be computed by max-product BP. We address both these concerns in the next subsections.

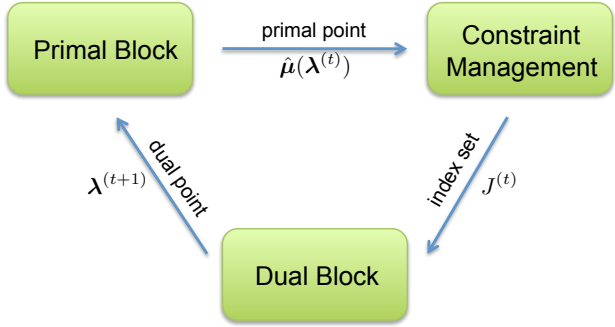

Figure 1: Overview of a Dynamic Supergradient Method. Figure adapted from [4].

### 6.1 Optimizing over Exponentially Many Dual Variables with a Dynamic Supergradient Method

In order to optimize over the exponentially large set of dual variables, we follow a *dynamic* supergradient method (DSM) (see [4] for an overview). Intuitively, DSMs can be thought of as the dual-procedure to the cutting-plane algorithm.

A DSM maintains an index set of *active* dual variables ($J = \cup_m J^m$, where $J^m \subset \{1, \ldots, |T(G)|\}$), and all inactive dual variables are fixed to zero, *i.e.* $\lambda_{mj} = 0, \forall j \notin J^m$. As visualized in Fig. 1, DSMs consist of three kinds of operations:

1. **Primal Block:** Given the current dual variable $\boldsymbol{\lambda}^{(t)}$, the primal block evaluates the Lagrangian (10) to find a new primal point $\hat{\boldsymbol{\mu}}(\boldsymbol{\lambda}^{(t)})$.

2. **Constraint Management Block:** Given the current primal point $\hat{\boldsymbol{\mu}}(\boldsymbol{\lambda}^{(t)})$, the constraint management block augments the index set of active dual variables to get the new index set $J^{(t)}$. Optionally, this block may also choose to drop some dual variables from the index set. This block is described in detail below.

3. **Dual Block:** Given the current index $J^{(t)}$, the dual block constructs the dual update direction (the supergradient) and stepsize to produce a new dual variable $\boldsymbol{\lambda}^{(t+1)}$ with $\lambda_{mj}^{(t+1)} = 0, \forall j \notin J^{m(t)}$.

The primal and dual blocks will be discussed in the next subsection, where we describe how the supergradient may be efficiently computed for general graphs. We now describe the constraint management block in detail. To develop an intuition for this step, recall that the complementary slackness condition [2] tells us that given a pair of optimal primal dual variables ($\boldsymbol{\mu}^*$ and $\boldsymbol{\lambda}^*$): $I^T(\boldsymbol{\mu}^*, \boldsymbol{\mu}^m) < 0 \implies \lambda_{mT}^* = 0$. Thus, intuitively the active set $J^{(t)}$ must focus on the dual variables corresponding to the violated inequalities. We use a maximum violated oracle that adds to the active set $J^{m(t)}$, the index of the dual variable corresponding to the most violated inequality, *i.e.*

$$J^{m(t+1)} \longleftarrow J^{m(t)} \cup \text{index}(\hat{T}) \quad (11)$$

where,

$$\hat{T} = \underset{T \in \mathcal{T}(G)}{\text{argmax}} \sum_{(i,j) \in T} w_{ij}, \quad \textbf{[Max-Wt Span-Tree]} \quad (12a)$$

$$w_{ij} = \hat{\boldsymbol{\mu}}_{ij}(\boldsymbol{\lambda}^{(t)}) \cdot \boldsymbol{\mu}_{ij}^m$$
$$- \hat{\boldsymbol{\mu}}_i(\boldsymbol{\lambda}^{(t)}) \cdot \boldsymbol{\mu}_i^m - \hat{\boldsymbol{\mu}}_j(\boldsymbol{\lambda}^{(t)}) \cdot \boldsymbol{\mu}_j^m. \quad (12b)$$

We note that this process is the dualized version of the cutting-plane method of Fromer and Globerson [7], where instead of adding the most violated spanning tree inequality to the LP, we include the index of its dual variable to the working set.

It can be shown that a dynamic supergradient method with such a maximum-violation-oracle constraint management block is guaranteed to converge to the optimum of the Lagrangian relaxation with the same choice of stepsize rules as standard supergradient methods. A rigorous proof can be found here [4]. DSMs actually allow for dual variables to be removed from the active set as well, under certain conditions. However, to keep the exposition simple and to match our implementation, we do not discuss that here, and refer the reader to [4].

The theory allows for several iterations of primal and dual blocks to be performed for each constraint management step. We found this to be crucial in practice.

### 6.2 Computing the Supergradient for General Graphs

Now let us address the problem of computing the supergradient, and implementing the primal and dual blocks. Given the current index set of active dual variables $J = \cup_m J^m$, the supergradient w.r.t. the active dual variables $\boldsymbol{\lambda}_J$ is given by:

$$\nabla_{\boldsymbol{\lambda}_{mj}} f(\boldsymbol{\lambda}) = I^j(\hat{\boldsymbol{\mu}}(\boldsymbol{\lambda}), \boldsymbol{\mu}^m) \quad \forall j \in J^m \quad (13)$$

where $\hat{\boldsymbol{\mu}}(\boldsymbol{\lambda})$ is an optimal primal solution of (10) for the current setting of $\boldsymbol{\lambda}$. For the case of tree-MRFs we could compute the optimal solution via dynamic programming. For general MRFs, the supergradient computation involves solving:

$$f(\boldsymbol{\lambda}) = \min_{\boldsymbol{\mu} \in \mathcal{L}(G)} \left\{ \sum_{i \in \mathcal{V}} \left( \boldsymbol{\theta}_i + \sum_{m=1}^{M-1} \sum_{j \in J^m} \lambda_{mj}(1 - d_i^{T_j}) \boldsymbol{\mu}_i^m \right) \cdot \boldsymbol{\mu}_i \right.$$
$$\left. + \sum_{(i,j) \in \mathcal{E}} \left( \boldsymbol{\theta}_{ij} + \sum_{m=1}^{M-1} \sum_{j \in J^m} \lambda_{mj} \boldsymbol{\mu}_{ij}^m \right) \cdot \boldsymbol{\mu}_{ij} \right\}, \quad (14)$$

where $G$ is now a graph with loops. Evaluating the above Lagrangian involves solving a MAP inference LP (with modified potentials) and thus any message-passing algorithm for MAP (MPLP [8], Dual-Decomposition [13], or Max-Sum Diffusion [26]) may be used to solve this problem. However, unlike the two-pass max-product BP used for tree-MRFs, these message-passing algorithms typically require many hundreds of iterations to converge. Running these iterations for each step of supergradient ascent can (and in practice does) become prohibitively slow.

The key to speeding up this process is to realize that $f(\boldsymbol{\lambda})$ is a *partial* Lagrangian, and if evaluating it is difficult, then this suggests that some more constraints should be dualized till the partial Lagrangian becomes tractable. This is precisely what we do, using ideas from the dual-decomposition literature [9,13].

**Expanding the Partial Lagrangian.** In order to expand the partial Lagrangian, we will first try to identify tractable (tree-structured) subcomponents. The spanning-tree inequalities are already tree structured. $G$ is not, but we can convert it into a collection of tree-structured factors.

Let $TC(G) = \{T_1, \ldots, T_p\}$ be a *spanning-tree cover* of $G$, i.e. a collection of spanning trees such that each edge of G appears in at least one tree in $TC(G)$. Our approach doesn't really need the trees to be spanning, but we describe the following with a spanning-tree cover to keep the notation simple. With a slight abuse of notation, we use $TC(i,j)$ to denote the subset of trees that contain edge $(i,j) \in \mathcal{E}$. Moreover, we decompose the original energy function $\boldsymbol{\theta}$ into a collection of energy functions $\{\boldsymbol{\theta}^T \mid T \in TC(G)\}$, one for each tree in the tree cover, such that:

$$\sum_{T \in TC(G)} \boldsymbol{\theta}_i^T = \boldsymbol{\theta}_i \ \forall i \in \mathcal{V} \ \& \ \sum_{T \in TC(i,j)} \boldsymbol{\theta}_{ij}^T = \boldsymbol{\theta}_{ij} \ \forall (i,j) \in \mathcal{E} \tag{15a}$$

$$\implies \sum_{T \in TC(G)} \boldsymbol{\theta}^T \cdot \boldsymbol{\mu} = \boldsymbol{\theta} \cdot \boldsymbol{\mu} \tag{15b}$$

This can be easily satisfied by distributing the node and edge energies "evenly", i.e. $\boldsymbol{\theta}_i^T = \frac{1}{|T(G)|}\boldsymbol{\theta}_i$, $\boldsymbol{\theta}_{i,j}^T = \frac{1}{|T(i,j)|}\boldsymbol{\theta}_{ij}$. Thus, these energies specify a tree decomposition [24] of $\boldsymbol{\theta}$.

Let us now assign to each tree in $TC(G)$, its own copy of the primal variables $\boldsymbol{\mu}^T, \forall T \in TC(G)\}$. Also assign to each spanning tree inequality in the active set $J^m$, its own copy of the primal variables $\boldsymbol{\mu}^{mj}, \forall j \in J^m$. Finally, we use $\boldsymbol{\theta}_i^{mj} \triangleq \lambda_{mj}(1 - d_i)\boldsymbol{\mu}_i^m$ to denote the node energy of the spanning tree factor $j \in J^m$, and $\boldsymbol{\theta}_i^{mj} = \lambda_{mj}\boldsymbol{\mu}_{ij}^m$ to denote the edge energy.

With these new variables, we can now write the existing partial Lagrangian as:

$$f(\boldsymbol{\lambda}) = \min_{\boldsymbol{\mu} \in \mathcal{L}(G), \tilde{\boldsymbol{\mu}}} \left\{ \sum_{T \in TC(G)} \boldsymbol{\theta}^T \cdot \boldsymbol{\mu}^T + \sum_{m=1}^{M-1} \sum_{j \in J^m} \boldsymbol{\theta}^{mj} \cdot \boldsymbol{\mu}^{mj} \right\} \tag{16a}$$

$$\text{s.t.} \quad \boldsymbol{\mu}_i^{\tau} = \tilde{\boldsymbol{\mu}}_i \quad \forall \tau \in TC(G) \cup J, \ \forall i \in \mathcal{V} \tag{16b}$$

This formulation of the partial Lagrangian uses a global variable $\tilde{\boldsymbol{\mu}}$ to force all tree-structured subproblems to agree on the labellings at the nodes, and thus is equivalent to the earlier formulation (14). However, we can now expand this partial Lagrangian by further dualizing these constraints (16b):

$$f(\boldsymbol{\lambda}, \boldsymbol{\delta}) = \min_{\boldsymbol{\mu}^{\tau} \in \mathcal{L}(\tau), \tilde{\boldsymbol{\mu}}} \left\{ \sum_{T \in TC(G)} \boldsymbol{\theta}^T \cdot \boldsymbol{\mu}^T + \sum_{m=1}^{M-1} \sum_{j \in J^m} \boldsymbol{\theta}^{mj} \cdot \boldsymbol{\mu}^{mj} \right.$$
$$\left. + \sum_{\tau \in TC(G) \cup J} \sum_{i \in \mathcal{V}} \boldsymbol{\delta}_{\tau i} \cdot \left(\boldsymbol{\mu}_i^{\tau} - \tilde{\boldsymbol{\mu}}_i\right) \right\} \tag{17}$$

where $\boldsymbol{\delta} = \{\boldsymbol{\delta}_{\tau i} \mid \forall \tau \in TC(G) \cup J, \ \forall i \in \mathcal{V}\}$ is the set of Lagrangian multipliers for the dualized equality constraints, and $\mathcal{L}(\tau)$ is now the local polytope of each of the tree-structured subproblems.

Notice that this expanded partial Lagrangian completely decouples into independent minimizations over tree-structured subproblems:

$$f(\boldsymbol{\lambda}, \boldsymbol{\delta}) = \sum_{\tau \in TC(G) \cup J} \min_{\boldsymbol{\mu}^{\tau} \in \mathcal{L}(\tau)} \left\{ \sum_{i \in \mathcal{V}} \left(\boldsymbol{\theta}_i^{\tau} + \boldsymbol{\delta}_{\tau i}\right) \cdot \boldsymbol{\mu}_i^{\tau} \right.$$
$$\left. + \sum_{(i,j) \in \mathcal{E}} \boldsymbol{\theta}_{ij}^{\tau} \cdot \boldsymbol{\mu}_{ij}^{\tau} \right\}$$
$$- \min_{\tilde{\boldsymbol{\mu}}} \left( \sum_{\tau \in TC(G) \cup J} \sum_{i \in \mathcal{V}} \boldsymbol{\delta}_{\tau i} \right) \cdot \tilde{\boldsymbol{\mu}}_i \tag{18}$$

We can see that the unconstrained minimization over $\tilde{\boldsymbol{\mu}}$ forces a constraint on the Lagrangian variables, i.e. $\sum_{\tau \in TC(G) \cup J} \sum_{i \in \mathcal{V}} \boldsymbol{\delta}_{\tau i} = 0$; otherwise the Lagrangian will not have finite value. We denote by $\Delta \triangleq \{\boldsymbol{\delta} \mid \sum_{\tau \in TC(G) \cup J} \sum_{i \in \mathcal{V}} \boldsymbol{\delta}_{\tau i} = 0\}$, the set of all feasible Lagrangian multipliers $\boldsymbol{\delta}$.

Note that the expanded partial Lagrangian can be efficiently evaluated by running two-pass max-product BP on each of the tree-structured subproblems. The number of such tree-structured subproblems is equal to $|TC(G)|+|J|$, i.e. the size of the spanning-tree cover (upper bounded by $n$, and typically a small constant) and the number of active spanning tree inequalities.

**Optimizing the Expanded Partial Lagrangian.** The dual problem for the expanded partial Lagrangian is given by $\max_{\boldsymbol{\lambda} \geq 0, \boldsymbol{\delta} \in \Delta} f(\boldsymbol{\lambda}, \boldsymbol{\delta})$. In the previous section, we described how the dual of the partial Lagrangian $f(\boldsymbol{\lambda})$ can be optimized with a dynamic supergradient method. Optimizing the dual of the expanded Lagrangian is very similar to the described procedure, with the minor modification that the dual variable $\boldsymbol{\delta}$ always stays in the active set $J$.

The supergradient w.r.t. $\boldsymbol{\delta}$ is given by:

$$\nabla_{\boldsymbol{\delta}_{\tau i}} f(\boldsymbol{\lambda}, \boldsymbol{\delta}) = \hat{\boldsymbol{\mu}}_i^{\tau}(\boldsymbol{\lambda}, \boldsymbol{\delta}) \quad \forall \tau \in TC(G) \cup J \ \forall i \in \mathcal{V}, \tag{19}$$

where $\hat{\boldsymbol{\mu}}_i^{\tau}(\boldsymbol{\lambda}, \boldsymbol{\delta})$ is an optimal primal solution of the tree suproblem $\tau$ for the current setting of $\boldsymbol{\lambda}$ and $\boldsymbol{\delta}$.

The projection step onto $\Delta$ is fairly simple – it involves satisfying the zero-sum constraint, which can be enforced by subtracting the mean of the dual variables. Overall, the dual update w.r.t. $\boldsymbol{\delta}$ is given by $\boldsymbol{\delta}^{(t+1)} = \left[\boldsymbol{\delta}^{(t)} + \alpha_t \nabla_{\boldsymbol{\delta}} f(\boldsymbol{\lambda}^{(t)}, \boldsymbol{\delta}^{(t)})\right]_0$, where $\alpha_t$ is the stepsize and $[\cdot]_0$ is the zero-projection operator *i.e.* $[\boldsymbol{\delta}^{(t)}]_0 = \{\boldsymbol{\delta}^{(t)}_{\tau i} - s \mid \forall \tau \in TC(G) \cup J, \forall i \in \mathcal{V}\}$, where $s = \frac{1}{|\boldsymbol{\delta}|} \sum_{\tau' \in TC(G) \cup J} \sum_{j \in \mathcal{V}} \boldsymbol{\delta}^{(t)}_{\tau' j}$.

The entire algorithm is summarized in Algorithm 1. We call our algorithm STEELARS for Spanning TrEe inEquality LAgrangian Relaxation Scheme.

---

**Algorithm 1** STEELARS

1: $(\boldsymbol{\mu}^*, \{\boldsymbol{\lambda}^*, \boldsymbol{\delta}^*\}) = \text{STEELARS}(G, \theta, \{\boldsymbol{\mu}^m\})$
2: **Input:** $G = (\mathcal{V}, \mathcal{E})$, graph instance,
3:         $\boldsymbol{\theta} = \{\theta_A \mid A \in \mathcal{V} \cup \mathcal{E}\}$ energy vector
4:         $\{\boldsymbol{\mu}^m \mid m \in [M-1]\}$, $M-1$ previous solutions
5: **Output:**   $\boldsymbol{\mu}^*_A \in [0,1]^{X_A}$, optimum solution vector to M-Best MAP LP
6:         $\{\boldsymbol{\lambda}^*, \boldsymbol{\delta}^*\}$, optimum dual variables
7: **Algorithm:**
8: Construct a tree-decomposition of G: $TC(G), \{\boldsymbol{\theta}^T \mid T \in TC(G)\}$
9: $\boldsymbol{\lambda}^{(0)} \longleftarrow \mathbf{0}; \boldsymbol{\delta}^{(0)} \longleftarrow \mathbf{0}; \hat{\boldsymbol{\mu}} \longleftarrow \mathbf{0}$
10: $J^{m(0)} \longleftarrow \emptyset, \quad \forall m \in [M-1]$
11: $t \longleftarrow 0$
12: **while** Not Converged **do**
13:     **{Constraint Management Block}**
14:     **for** $m = 1, \ldots, M-1$ **do**
15:         $w_{ij} = \hat{\boldsymbol{\mu}}_{ij} \cdot \boldsymbol{\mu}^m_{ij} - \hat{\boldsymbol{\mu}}_i \cdot \boldsymbol{\mu}^m_i - \hat{\boldsymbol{\mu}}_j \cdot \boldsymbol{\mu}^m_j$
16:         $\hat{T} = \arg\max_{T \in \mathcal{T}(G)} \sum_{(i,j) \in T} w_{ij}$ {Max-weight Spanning Tree.}
17:         $J^{m(t+1)} \longleftarrow J^{m(t)} \cup \text{index}(\hat{T})$
18:     **end for**
19:     $J^{(t)} \longleftarrow \cup_{m \in [M-1]} J^{m(t)}$
20:     {Multiple Iteration of Primal & Dual Blocks}
21:     **for** $t' = 1, \ldots, 20$ **do**
22:         $\boldsymbol{\theta}^{mj}_i \longleftarrow \lambda^{(t)}_{mj}(1 - d_i)\boldsymbol{\mu}^m_i$
            $\boldsymbol{\theta}^{mj}_{ij} \longleftarrow \lambda^{(t)}_{mj}\boldsymbol{\mu}^m_{ij} \quad \forall m \in [M-1], \forall j \in J^m$
23:         **{Primal Block}**
24:         **for** $\tau \in TC(G) \cup J$ **do**
25:             $\boldsymbol{\theta}^\tau_i \longleftarrow \boldsymbol{\theta}^\tau_i + \boldsymbol{\delta}^{(t)}_{\tau i}; \boldsymbol{\theta}^\tau_{ij} \longleftarrow \boldsymbol{\theta}^\tau_{ij}$
26:             $\hat{\boldsymbol{\mu}}^\tau(\boldsymbol{\lambda}^{(t)}, \boldsymbol{\delta}^{(t)}) = \arg\min_{\boldsymbol{\mu}^\tau \in \mathcal{L}(\tau)} \boldsymbol{\theta}^\tau \cdot \boldsymbol{\mu}^\tau$ {Two-pass max-product BP.}
27:         **end for**
28:         **{Dual Block}**
29:         $\nabla_{\lambda_{mj}} f(\boldsymbol{\lambda}^{(t)}, \boldsymbol{\delta}^{(t)}) \longleftarrow I^j(\hat{\boldsymbol{\mu}}^{mj}(\boldsymbol{\lambda}^{(t)}, \boldsymbol{\delta}^{(t)}), \boldsymbol{\mu}^m)$
30:         $\nabla_{\delta_{\tau i}} f(\boldsymbol{\lambda}^{(t)}, \boldsymbol{\delta}^{(t)}) \longleftarrow \hat{\boldsymbol{\mu}}^\tau_i(\boldsymbol{\lambda}^{(t)}, \boldsymbol{\delta}^{(t)})$
31:         $\boldsymbol{\lambda}^{(t+1)} \longleftarrow \left[\boldsymbol{\lambda}^{(t)} + \alpha_t \nabla_{\boldsymbol{\lambda}} f(\boldsymbol{\lambda}^{(t)}, \boldsymbol{\delta}^{(t)})\right]_+$
32:         $\boldsymbol{\delta}^{(t+1)} \longleftarrow \left[\boldsymbol{\delta}^{(t)} + \alpha_t \nabla_{\boldsymbol{\delta}} f(\boldsymbol{\lambda}^{(t)}, \boldsymbol{\delta}^{(t)})\right]_0$
33:         $t \longleftarrow t + 1$
34:     **end for**
35:     $\hat{\boldsymbol{\mu}} \longleftarrow$ Best Feasible Primal So Far.
36:     $\hat{\boldsymbol{\mu}}^* \longleftarrow$ Running Average of $\hat{\boldsymbol{\mu}}^\tau$
37: **end while**

---

**Tightness of STEELARS.** At first glance, it may seem like the expanded partial Lagrangian (18) solves a weaker relaxation than our original partial Lagrangian (10). However, similar to our argument in the tree-MRF case, problem (9) is an LP. Thus, strong duality holds and *all* partial Lagrangians achieve the same optimum. This implies that even for non-tree MRFs, STEELARS *exactly* solves the M-Best MAP LP of Fromer and Globerson [7] and does not introduce any new approximation gap to the integer program. Moreover, this guarantee this does not depend on the choice of the spanning-tree cover $TC(G)$; in fact, any tree cover (even non-spanning) may be used.

Finally, note that we described the expanded partial Lagrangian in terms of tree-structured subproblems. However, any efficient subproblem may be used, *e.g.* submodular subproblems solved with graph-cuts [12].

## 7 Experiments

**Setup.** We tested our algorithm in three scenarios:

1. Tree MRFs with $M = 2$. This is the simplest case, where the M-Best MAP LP is guaranteed to be tight. Moreover, there is a single spanning tree inequality and the constraint management block plays no role.

2. 2-label Submodular MRFs with $M = 5$. For such problems the MAP LP is guaranteed to be tight, but not the M-Best MAP LP. Moreover, a tree decomposition is not required because the 2-label submodular factor may be efficiently minimized via graph-cuts [12].

3. General 4-label loopy MRFs with $M = 5$. This is the most general case described in Section 6.

**Baselines.** We compared our algorithm with the STRIPES algorithm of Fromer and Globerson [7], the Lawler-Nilsson algorithm [16,18], and the BMMF algorithm of Yanover and Weiss [28]. Recall that STRIPES does not directly solve the M-Best MAP LP, but rather solves a sequence of 2nd-Best MAP LPs encapsulated in the Lawler-Nilsson [16, 18] partitioning scheme. This is a tradeoff between efficiency and accuracy – it would be more efficient to directly solve the M-Best MAP LP, but the LP is fractional and thus the partitioning scheme performs better. Note that STEELARS could also be encapsulated in the Lawler-Nilsson [16, 18] partitioning scheme in a straightforward manner. However, we wish to study the performance of the Lagrangian relaxation for the M-Best MAP LP directly, and leave this partitioning scheme extension for future work.

**Implementation details.** The implementations for STRIPES and Lawler-Nilsson were provided by the authors of [7], while BMMF is provided by the authors of [28]. For STRIPES, the LPs in each iteration were solved using the `GNU LPK` library. STEELARS is implemented in MATLAB, but max-product BP is written in C++ for efficiency. All experiments are performed on a 64-bit 8-Core Intel i7 machine with 12GB RAM and the timing reported is cputime. Following [15], we chose the stepsize at iteration $t$ to be $\alpha_t = \frac{1}{\eta_t+1}$, where $\eta_t$ is the number of times the objective value $f(\boldsymbol{\lambda}^{(t)}, \boldsymbol{\delta}^{(t)})$ has decreased from one iteration to the next. This rule has the same convergence guarantees as the standard $\alpha_t = \frac{1}{t}$ decaying rule, but empirically performs much better.

**Integer Primal Extraction.** STEELARS is a dual-ascent algorithm and thus always maintains a feasible dual solution, but not necessarily an (integer) primal feasible solution, which is what we are interested in. However, since the primal block is repeatedly called for computing the supergradient, we simply keep track of the best (integer) primal feasible solution produced so far, and output that at the end of the algorithm.

**Evaluation.** We compare different algorithms on two metrics – run-time and accuracy of solutions returned. For tree-MRFs with $M = 2$ all methods are guaranteed to return exact solutions, and thus we can simply compare the run-times. For general MRFs, we follow the protocol of [7] and measure relative accuracy of different methods. Specifically, we pool all solutions returned by all methods, note the top $M$ solutions in this pool, and then for each method report the fraction of these solutions that it contributed.

Our results will demonstrate the effectiveness of STEELARS primarily in terms of efficiency. We will show that since STEELARS is a message-passing algorithm it is significantly faster than a generic LP solver (`GLPK`), sometimes by *orders of magnitude* even though it is guaranteed to converge to the same solution. STEELARS naturally scales to large instances that were previously unsolvable using LP solvers.

**Tree MRFs with $M = 2$.** We generated synthetic problems by sampling random spanning trees on $n$ nodes. Each variable could take 4 labels. Node and edge potentials were sampled from standard Gaussians. For $M = 2$ the M-Best MAP LP is guaranteed to be tight, and thus both STRIPES and STEELARS produces precisely the same answers. Fig. 2a shows the time taken by both algorithms as a function of the size of the tree $n$, averaged over 5 samplings of the parameters. Note that the x-axis is in log-scale. As $n$ increase, STRIPES very quickly becomes intractable, with `GPLK` ultimately running out of memory. However, STEELARS shows a much better behaviour in runtime. Fig. 2 also shows the value of the dual and best feasible integer primal produced by STEELARS as a function of the number of iterations. Recall that for STEELARS each iteration corresponds to running max-product BP on a single tree. Fig. 2d shows that generally the number of such iterations is pretty low $\sim 12$. Note that since this is a tree MRF, BMMF would only require 2 call to BP to compute max-marginals under its partitioning scheme (assuming it was carefully implemented to recognize a tree-graph and not run asynchronous BP). Thus STEELARS is not too much slower than the optimal thing to do for a tree. Of course, BMMF would also require multiple iterations of synchronous BP on loopy graphs, which is what we checked next.

**2-label Submodular MRFs.** For this scenario, we constructed $\sqrt{n} \times \sqrt{n}$ grids. Each variable could take 2 labels. We again sampled node and edge energies from Gaussians, but ensured that edge energies were submodular. This allows for the use of graph-cuts [12] for optimizing the submodular factor. Fig. 3 shows the value of the dual and best feasible integer primal as a function of the number of iterations. The sharp falls in the dual correspond to the constraint management block calling the separation oracle to increment the working set. Each iteration now involves one call to graph-cuts and $|J^{(t)}|$ calls to two-pass max-product BP. Notice that the number of iterations are much larger than for the tree MRF ($\sim$150 as opposed to 12). However, max-flow algorithms in general and the implementation of [12] in particular, are highly efficient. Fig. 3a shows the run-time of all algorithms as a function of $n$. We can see that STEELARS is by far the fastest, with other algorithms becoming intractable very quickly. Unfortunately, as Fig. 3d shows it is also the least accurate, validating the choice of Fromer and Globerson [7] to not solve the LP directly, and use a partitioning scheme instead. We plan to follow up on this direction.

Interestingly, Fig. 3a seems to suggest that BMMF performs worse than STRIPES, even though BMMF simply involves M calls to loopy BP. We believe this an artifact caused by the fact that the BMMF implementation of [28] is written in MATLAB, and not particularly optimized. Moreover, in this specific experiment, it could be made faster by computing max-marginals via the approach of [11] instead of loopy BP. At a high-level, the key difference between BMMF and STEELARS is analogous to the difference between loopy BP and MPLP – both are message-passing algorithms, but only one solves the LP relaxation and provides improving lower-bounds.

**General MRFs.** For this scenario, we constructed $\sqrt{n} \times \sqrt{n}$ grid graphs as well, but each variable could take 4-labels and edge energies were not restricted to be "attractive". We used a standard two-tree decomposition of the grid graph. Thus, each iteration of supergradient ascent involved $2 + |J^{(t)}|$ calls to max-product BP. We observed trends similar to the submodular MRFs case, both in terms of number of iterations required for STEELARS to converge, and in the relative standing w.r.t. the baselines. This case is specially interesting because for small values of $|J^{(t)}|$, the work done by STEELARS (*i.e.* ∼150 iteration of two-pass BP) for solving M-Best MAP is comparable to running TRW [13, 24] for MAP. Moreover, we could run STEELARS on a $300 \times 500$ image labelling problem, but similar to the observation of [27], could not even solve the MAP LP with `GLPK`.

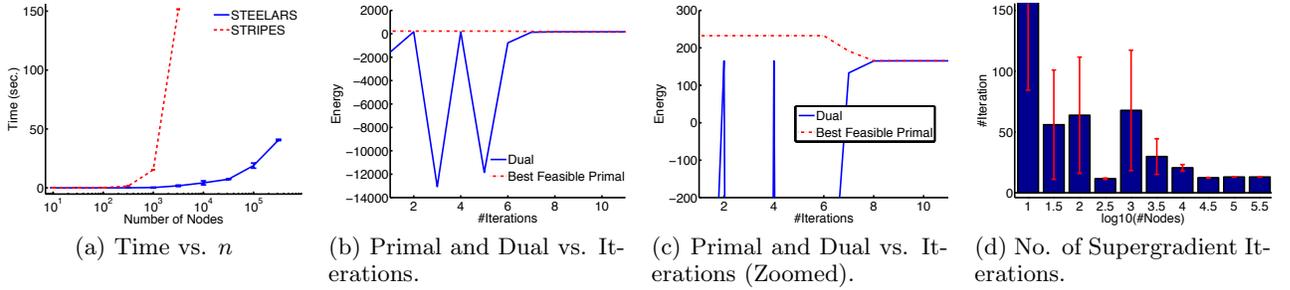

Figure 2: Tree MRF with $M = 2$.

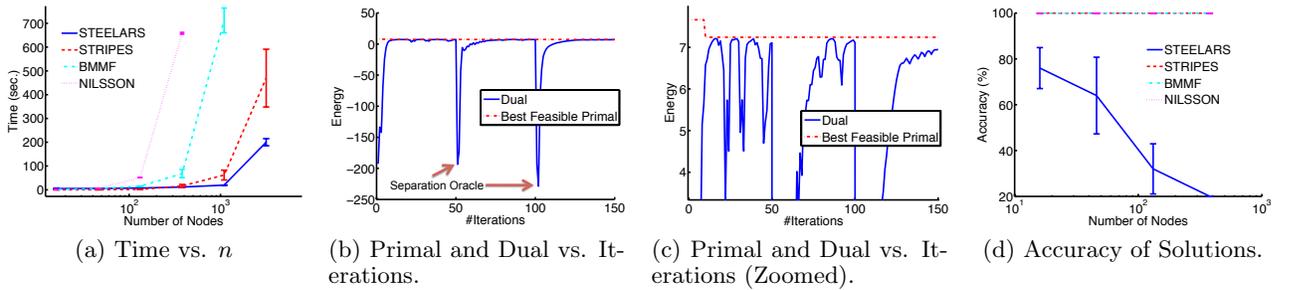

Figure 3: 2-label Submodular MRFs with $M = 5$.

## 8 Conclusions

In conclusion, we presented the *first* message-passing algorithm for solving the LP relaxation of the M-Best MAP problem in discrete undirected graphical models. Our approach used a particular Lagrangian relaxation to construct a partial Lagrangian that allowed the use of combinatorial optimization algorithms. To handle the exponentially large set of constraints, we used a dynamic supergradient scheme that is essentially a dual procedure to the cutting-plane algorithm. Our message-passing algorithm retains all the guarantees of the LP formulation of Fromer and Globerson [7], while being *orders of magnitude* faster.

**Extracting Diverse M-Best Solutions.** In a number of applications, especially computer vision, the M-Best MAP solutions are essentially minor perturbations of each other. In concurrent work [1], we have also presented a solution to the *Diverse* M-Best MAP problem, where given a measure of 'distance' between two solutions, we block all solutions within some $k$-distance-ball of the previous solutions.

We hope that both these algorithms will be useful for practitioners where the problem size prohibits the use of generic LP-solvers.

**Future Work.** There are a number of interesting directions in front of us. In the short-term, we are interested in encapsulating STEELARS inside the Lawler-Nilsson partitioning scheme, similar to STRIPES to further increase the accuracy of the method. Since the M-Best MAP LP is so often fractional, another direction is to *tighten* the LP, *e.g.* using techniques proposed by Sontag *et al.* [23]. It would also be interesting to compare LP-relaxation based methods like STRIPES and STEELARS with heurisitic-search methods [3].

**Acknowledgements.** We thank Sebastian Nowozin for being the human encyclopedia of optimization and pointing us to the literature of Dynamic Subgradient Methods; Amir Globerson for sharing the STRIPES implementation and useful discussions; Danny Tarlow for discussions that helped formalize the formulation.